# Wearable Sensors for Spatio-Temporal Grip Force Profiling

**Rongrong Liu [1]**, **Florent Nageotte [1]**, **Philippe Zanne[1]**, **Michel de Mathelin[1] and Birgitta Dresp-Langley[2]**

[1] ICube Lab, Robotics Department, University of Strasbourg, UMR 7357, CNRS, 67000 Strasbourg, France
[2] ICube Lab, UMR 7357, Centre National de la Recherche Scientifique CNRS, 67000 Strasbourg, France
E-mail: rong.rong.liu@unistra.fr; Nageotte@unistra.fr; zanne.philippe@unistra.fr; demathelin@unistra.fr; birgitta.dresp@unistra.fr;

**Summary:** Wearable biosensor systems with transmitting capabilities represent innovative technology developed to monitor exercise and other task activities. This technology enables real-time, convenient, and continuous monitoring of a user's behavioral signals, relative to body motion, body temperature and a variety of biological or biochemical markers, like individual grip force, which is studied in this paper. To achieve this goal, a four-step pick-and-drop image-guided robot-assisted precision task has been designed using a wearable wireless sensor glove system. The spatio-temporal grip force profiling is analyzed on the basis of thousands of individual sensor data collected from the twelve locations on the dominant and non-dominant hands of each of the three users in ten successive task sessions. Statistical comparison have shown specific differences between the grip force profiles of individual users as a function of task skill level and expertise. image-guided robot-assisted precision task

## 1. Introduction

Wearble sensors, as the name implies, are integrated into wearable objects or directly with the body in order to monitor and transmit a user's behavioral signals in real time. In this paper, the spatio-temporal grip force profiling will be studied based on the data collected from a wearable wireless sensor glove system developed in the lab [1] [2].

## 2. Materials and Methods

A specific wearable sensor system in terms of a glove for each hand with inbuilt Force Sensitive Resistors (FSR) has been developed. The hardware and software configurations will be briefly described here below. One may go to https://www.mdpi.com/1424-8220/19/20/4575/htm for further detailed information.

The wireless sensor glove hardware-software system, is designed for bi-manual intervention, and task simulations may solicit either the dominant or the non-dominant hand, or both hands at the same time. For each hand, twelve anatomically relevant FSR are emplyed to measure the grip force aplliled on certain locations on the fingers and in the palm. These FSR have been sewn into a soft glove (Fig. 1 (a)) and their locations are shown in Fig. 1 (b).

The software of the glove system includes two parts: one running on the gloves, and the other running on the computer algorithm for data collection. During the experiment, each of the two gloves is sending data to the computer separately every 20 milliseconds (50Hz), merged with the time stamps and sensor identification. This data package is sent to the computer via Bluetooth, which will be then decoded and saved by the computer software.

A four-step pick-and-drop image-guided robot-assisted precision task, as described in Table 1, has been designed for this individual grip force study. Grip force data are analyzed here for one left-handed highly proficient expert, and one right-handed complete novice.

## 3. Results

Several thousands of grip force data have been collected from the twelve sensor locations in ten successive sessions for repeated execution of the pick-and-drop robotic task. The corresponding task times for the dominant and non-dominant hands of each of the three users are illustrated in Table 2.

As the middle phalanx of the small finger allows for precision grip force control [3] [4], and is critically important in surgical and other precision tasks, we focus on the corresponding sensor S7 on the dominant hand of expert and novice as the most representative.

Individual spatio-temporal grip force profiles have been plotted in Fig. 2 for the first and the last individual sessions, in terms of average peak amplitudes (mV) for fixed successive temporal windows of 2000 milliseconds (100 signals per time window, session, and user).

A 2-Way ANOVA on the raw grip force data has been conducted for statistical comparison. The expertise-specific difference between the two user profiles is characterized by the novice deploying largely insufficient grip forces, from the first session with m=98mV /sem=1.2 to the last with m=78mV/sem=1.6, while the expert produces sufficient grip force for fine movement control from the first session with m=594mV /sem=1.8 to the last with m=609mV/sem=2.2. The interaction between the 'expertise' and 'session' factors for sensor S7 is highly significant with F( 1,2880)=188.53; p<.001.

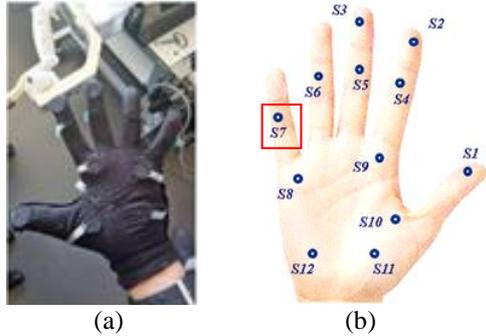

(a)        (b)

**Fig. 1.** Signals relative to grip force are sampled from 12 anatomically relevant FSR locations on the fingers and in the palm of both hands.

**Table 1.** Four-step pick-and-drop task

| Step | Description |
|---|---|
| 1 | Activate and move tool towards object location |
| 2 | Open and close grippers to grasp and lift object |
| 3 | Move tool with object to target location |
| 4 | Open grippers to drop object in box |

**Table 2.** Task time (sec)

| User | Dominant | Non-dominant |
|---|---|---|
| Expert | 8.88 | 10.19 |
| Trained | 11.90 | 13.53 |
| Novice | 15.42 | 12.99 |

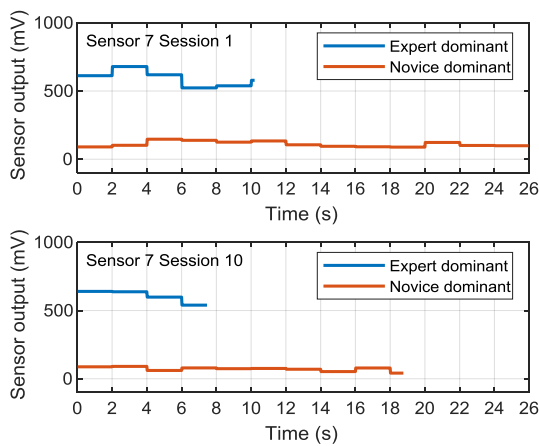

**Fig. 2.** Individual spatio-temporal grip force profiles showing average peak amplitudes (mV) from sensor S7 for fixed successive temporal windows of 2000 milliseconds for the first and last of ten sessions of two users.

## 4. Discussion

The spatio-temporal profiling and statistical comparison have shown specific differences between the grip force profiles of individual users as a function of task skill level and expertise in using the robotic system. Experts and non-experts employ different grip-force strategies, reflected by differences in amount of grip force deployed by the middle phalanx of the small finger, with the novice dominant hand deploying insufficient grip forces, and no major evolution between the first and the last task sessions.

In terms of task time, at the beginning, the novice takes more than twice as long performing the precision task by comparison with the expert, but at the end scores a 30% time gain indicating a considerable temporal training effect.

## 5. Conclusions

Grip force analysis on wearable sensors signals is a powerful means of tracking the evolution of individual force profile. The analyses shown in this paper here can deliver insight to monitor manual/bimanual precision tasks, control performance quality, or prevent risks in robot-assisted surgery systems, where excessive grip forces can cause tissue damage [5].

## Acknowledgements

The study is funded by the Excellence Programme of the University of Strasbourg. Material support by CNRS is gratefully acknowledged by the authors.